\documentclass[10pt,twocolumn,letterpaper]{article}

\usepackage[pagenumbers]{cvpr} % To force page numbers, e.g. for an arXiv version

\usepackage{algorithm}
\usepackage{algorithmic}
\usepackage{newfloat}
\usepackage{listings}
\usepackage{amsmath,amssymb}
\usepackage{threeparttable}
\usepackage{multirow}
\usepackage{booktabs}
\usepackage{nicefrac}
\usepackage{enumitem}
\usepackage{soul}

\def\Figref#1{Figure~\ref{#1}}
\def\Tableref#1{Table~\ref{#1}}
\newcommand{\figleft}{{\em (Left) }}

\newcommand{\figright}{{\em (Right) }}
\newcommand{\figtop}{{\em (Top) }}
\newcommand{\figbottom}{{\em (Bottom) }}
\newcommand{\CIS}{{CIS}}
\newcommand{\FullCIS}{{Components Inclusion Score}}
\newcommand{\FID}{{FID}}
\newcommand{\FullFID}{{Fréchet Inception Distance}}
\newcommand{\IS}{{IS}}
\newcommand{\FullIS}{{inception Score}}

\usepackage[dvipsnames]{xcolor}

\definecolor{cvprblue}{rgb}{0.21,0.49,0.74}
\usepackage[pagebackref,breaklinks,colorlinks,citecolor=cvprblue]{hyperref}

\def\paperID{9126} % *** Enter the Paper ID here
\def\confName{CVPR}
\def\confYear{2024}

\title{The Challenges of Image Generation Models \\ in Generating Multi-Component Images}

\author{Tham Yik Foong\\
Kyushu University\\
\and
Shashank Kotyan\\
Kyushu University\\
\and
Po Yuan Mao\\
Kyushu University\\
\and
Danilo Vasconcellos Vargas\\
Kyushu University\\
University of Tokyo\\
}

\begin{document}
\maketitle

\begin{abstract}

Recent advances in text-to-image generators have led to substantial capabilities in image generation. 
However, the complexity of prompts acts as a bottleneck in the quality of images generated.
A particular under-explored facet is the ability of generative models to create high-quality images comprising multiple components given as a prior.
In this paper, we propose and validate a metric called \FullCIS~(\CIS) to evaluate the extent to which a model can correctly generate multiple components. 
Our results reveal that the evaluated models struggle to incorporate all the visual elements from prompts with multiple components ($8.53\%$ drop in \CIS~ per component for all evaluated models).
We also identify a significant decline in the quality of the images and context awareness within an image as the number of components increased ($15.91\%$ decrease in \FullIS~ and $9.62\%$ increase in \FullFID). 
To remedy this issue, we fine-tuned Stable Diffusion V2 on a custom-created test dataset with multiple components, outperforming its vanilla counterpart.
To conclude, these findings reveal a critical limitation in existing text-to-image generators, shedding light on the challenge of generating multiple components within a single image using a complex prompt.

\end{abstract}  

\section{Introduction}

\begin{figure}[!t]
\centering
\includegraphics[width=\columnwidth]{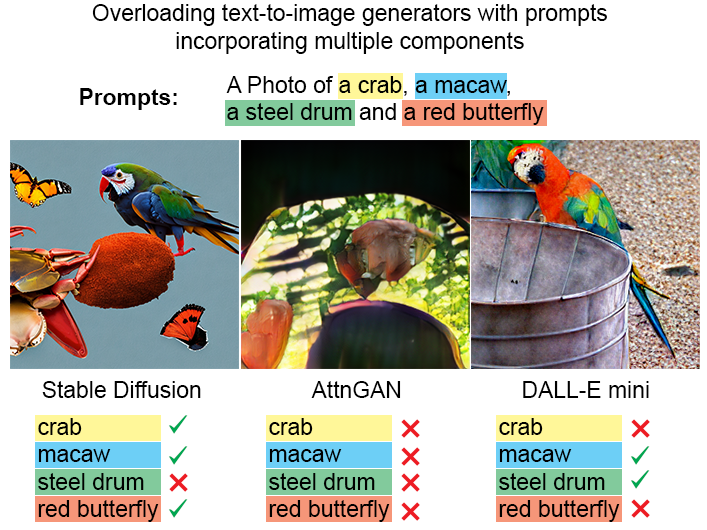}
\caption{
\textbf{Image generation models struggle to incorporate all the components in the generated image when given prompts involve several components.} 
Feeding image generation models with an example prompt \textit{`A photo of a crab, a macaw, a steel drum, and a red butterfly'}. Although the text instructs the creation of four components, the generated images illustrate that none of the example models (Stable Diffusion, AttnGan, DALL-E mini) can incorporate all four objects into a single image.
}
\label{introduction}
\end{figure}

Generating images with the help of neural networks is one of the challenging tasks in Computer Vision. 
There exist several architectures and methods based on either a) Variational Auto-Encoders (VAEs) like DCVAE \cite{parmar2021dual}, b) Generative Adversarial Networks (GANs) like, Attention GAN \cite{xu2018attngan}, Style GAN \cite{Karras2019stylegan2, sauer2022stylegan}, Big GAN \cite{brock2018large} or c) the more recent Diffusion-based Models like Dall-E \cite{ramesh2021zeroshot, dalle_mini, ramesh2022hierarchical}, Imagen \cite{saharia2022photorealistic}, Stable Diffusion \cite{rombach2021highresolution} to generate high-quality realistic images. 
Since the emergence of diffusion models, numerous methods have been further developed to improve the performance of diffusion models and extend their capacity to generate more diverse and high-fidelity images. 

% With the introduction of Diffusion-based Models, image generation through text prompts has reached new levels of realism.
% Generated images now have higher fidelity and diversity than those generated by GANs or VAEs. 
% Since their emergence, numerous methods have been developed to improve the performance of diffusion models and also extend the capacity of diffusion models to generate realistic images. 
% niche, gap, research question
% Problem definition/statement.
However, current image generation models perform impressively only when generating a single component with detailed instructions. 
They often struggle to incorporate all the components in the generated image when prompts involve several components \cite{hinz2018generating, he2021context, liu2022compositional}, implying that models are somewhat biased towards some parts of the prompt while ignoring the other parts. 
Moreover, there is a noticeable decline in both quality of the generated image and its context awareness with an increase in the complexity of the text prompt. 
This highlights the challenge of understanding the process of integrating multiple components within a single image, that even state-of-the-art image generators struggle with.
% Moreover, when models attempt to create images with numerous components, there is a noticeable decline in both the quality of the image and its context awareness.
% Among many challenges in image generation, the capability of these generative models to generate images with multiple components (objects of different classes) remains relatively underexplored.
% TODO perhaps not good/ maybe talk about robustness?

% TODO weak part

In the \Figref{introduction}, we can observe that when we overload the image generation models to create four distinct objects, none of the image generation models can fit all four objects into a single image.
The example prompt: \textit{`A photo of a crab, a macaw, a steel drum and a red butterfly'}, explicitly contained information to create four distinct objects. 
The fact that complex prompts limit the capability of current image generation models can be observed by overloading the prompts to include more than one component in the image.
This limitation can be further exploited by including more components in the prompt inducing a lower quality rendered image from state-of-the-art image generation models.

In this article, we rigorously test the current image generation models by overloading them with prompts incorporating multiple components and evaluate their capability of handling multiple components in the prompt. 
Specifically, our key contributions are listed below,

\noindent
\textbf{Multi Component Image Dataset (MCID):}
We introduce a test dataset called MCID, which contains a set number of components in a single image created by combining multiple images of the ImageNet dataset \cite{russakovsky2015imagenet}. 

\noindent
\textbf{Components Inclusion Score (CIS):}
We propose a novel \CIS~ metric to quantitatively measure a model's ability to incorporate multiple components from a prompt into the generated image. 

\noindent
\textbf{Image Generation Models fail to incorporate Multiple Components in Single Image:}
Our evaluation metric confirms a decrease of $8.53\%$ in \CIS~ per component, observed across prompts with up to $8$ components.
This also led to an overall decline in image quality, as reflected by the \FullIS~ and \FullFID~ metrics.
% Our results show that current image generation models struggle with visual realism as the number of components increases, often producing twisted, distorted, or incomplete components. 

\noindent
\textbf{Improve Multi-Component Generation Capability through Fine-Tuning:}
We improved the CIS of Stable Diffusion V2 by fine-tuning it on MCID, showcasing enhanced capability in multi-component image generation. This underscores the importance of expanding data distribution with images featuring multiple components.
%\footnote{For anonymity purposes, the code, MCID dataset, and fine-tuned weight will be released upon acceptance.}

% We evaluate these models using existing metrics like \FullFID~(\FID) and \FullIS~(\IS) to measure the quality of the image generated.
% In addition to the metric, we also curated the Multi-Component Image Dataset (MCID), composed of numerous multi-component images, to validate our metric and facilitate evaluation.
% This challenge largely stems from the nature of the training data these models learn from, which typically do not comprise images with numerous components. 
% Result
% Therefore, the evidence leads to the conclusion that modern image generators failed to generate images with multiple components.
% TODO contribution in bullets form
% The key contributions of this study can be summarized as follows: (1) We introduce the CIS metric to evaluate image generation models' capability to generate multi-component images quantitatively. (2) A dataset\footnote{For anonymity purposes, the code and dataset will be released upon acceptance.} of multi-component images available for public use. (3) Through experiments, we have verified the problem of inequality when current text-to-image generators generate images comprising multiple components.

\section{Related Work}

% FID score and IS Score are the two main metric which been widely used for evaluating generative models' performance. Both of these rely on a pre-existing classifier, Inceptionv3, which trained on ImageNet. However, their usage are difference. FID \cite{heusel2017gans} measures the Wasserstein-2 (Fréchet) distance between multivariate Gaussians fitted to the embedding space of the Inception-v3 network for both generated and real images. Technically, it is the 2048-dimensional activation vector of its pool3 layer. On the other hand, the IS score \cite{salimans2016improved} calculates the Kullback-Leibler divergence between the conditional class distribution and the marginal class distribution over the generated data. However, these two metric can only be used for measuring the quality of images. Difference from metric above, 

\subsection{Current Challenges}
% Specific Limitations in Image Generation
% Unresolved Minor Challenges in Image Generation

Despite the recent advances in text-to-image generators leading to outstanding image-generation capabilities, the internal workings of these generators remain largely unknown. 
For example, the regions between occluded objects are often poorly rendered due to a lack of context-aware modeling \cite{he2021context, ashual2019specifying, liao2021image}. 
Visual realism and coherence with the prompt also diminish when the prompt becomes too intricate or complex \cite{qiao2019learn, qiao2019mirrorgan}. 
An additional mystery is that generators can sometimes produce gibberish wording, interpretable only within the specific context of the generator itself \cite{milliere2022adversarial, daras2022discovering}. 
These phenomena can typically be evaluated subjectively rather than quantitatively.
Yet, the lack of evaluation measures for these specific issues hinders progress in fully resolving them.

\subsection{Evaluation Metrics}

\FullFID~(\FID) and \FullIS~(\IS) hold a prominent position as two of the most widely employed metrics in the domain of image generative models. 
The IS metric \cite{salimans2016improved} evaluates the diversity and quality of generated images by quantifying the discernibility of different classes within the generated dataset.
%In detail, it calculates the Kullback-Leibler divergence between the distribution of class labels for generated images and the overall class distribution across the generated dataset. 
%On a parallel note, the FID metric \cite{heusel2017gans} measures the dissimilarity between the statistical characteristics of both the generated images and real images.
%It quantifies their Wasserstein-2 distance between multivariate Gaussian distributions fitted to the embedding space of the Inception-v3 \cite{szegedy2016rethinking} network.
On a parallel note, The FID metric \cite{heusel2017gans} quantifies the dissimilarity between generated and real images by measuring the Wasserstein-2 distance between their multivariate Gaussian distributions fitted to the Inception-v3 \cite{szegedy2016rethinking} network's embedding space.

%These metrics constitute a significant part of the toolkit for assessing the performance of such models and are anchored in the utilization of a pre-existing classifier, typically one trained on the extensive ImageNet dataset.
%The IS metric, introduced by \cite{salimans2016improved}, takes a distinctive approach to evaluating the quality of generated images. 
%It calculates the Kullback-Leibler divergence between the conditional class distribution – which reflects the distribution of class labels for generated images – and the marginal class distribution, representing the overall class distribution across the generated dataset. 
%In essence, the IS metric offers insight into the diversity and quality of generated images by quantifying the discernibility of different classes within the generated dataset.

%On a parallel note, the FID metric, as presented by \cite{heusel2017gans}, offers another lens through which to assess the efficacy of image generative models. 
%Rather than focusing on the distribution of class labels, FID delves into the distribution of image features. 
%It quantifies the Wasserstein-2 distance between multivariate Gaussian distributions fitted to the embedding space of the Inception-v3 network for both the synthesized images and real images. 
%This effectively measures the dissimilarity between the statistical characteristics of the two sets of images, providing a holistic evaluation of their visual dissimilarity.

Both the IS, FID metrics, and their variants \cite{nash2021generating, liu2018improved, chong2020effectively} inherently concentrate on the visual fidelity and quality of the generated images. 
They offer valuable insights into the capabilities of generative models to capture the complexity of real-world images. 
In comparison, the essence of our metric diverges from this trajectory. 
Instead of assessing image quality per se, our metric centers on investigating the intricate interplay between multi-component prompts and the resultant images.

\begin{figure*}[!t]
\centering
\includegraphics[width=\textwidth]{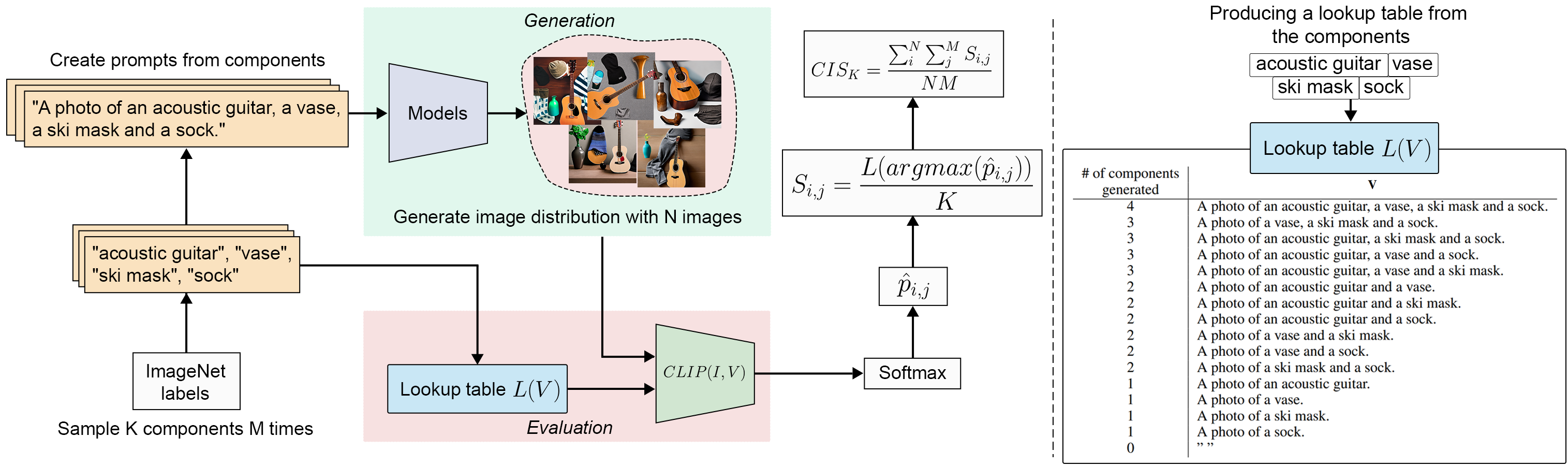}
\caption{
\textbf{The framework of CIS metric.} 
\figleft Multi-component prompts are constructed by sampling from the components pool (ImageNet labels), and the evaluated models generate image distribution based on these prompts. 
In the \textit{Evaluation} module, lookup tables \figright are created based on the sampled components. 
The CLIP model computes the softmax probability for each generated image $I$ corresponding to the prompts in $V$. 
An individual score $S$ is computed as a normalized sum of the successfully incorporated components from the prompt into the generated image. 
We determine the number of generated components by referencing the lookup table. Finally, the score $CIS_K$ for $K$ components is computed across all $S$.
}
\label{fig1}
\end{figure*}

Nevertheless, IS and FID has been criticized for certain limitations, such as being sensitive to class imbalance \cite{chong2020effectively}, not capturing image-to-image variations effectively, and inconsistency with human perception \cite{alaa2022faithful, binkowski2018gans, parmar2022aliased, ding2022cogview2}.
Therefore, the community tends to also employ other specialized metrics alongside the standard IS and FID.
For instance, the R-precision for text-to-image synthesis task \cite{xu2018attngan}, ad hoc network model as concept detector \cite{liu2022compositional}, caption
quality evaluation on image content \cite{jiang2019tiger, papineni2002bleu}, or even rely on human evaluation \cite{otani2023toward}.
These methods each have their own trade-off, and may specialize in certain evaluation tasks.

Among these methods, a closely related work CLIP Score \cite{hessel-etal-2021-clipscore} is a novel image captioning evaluation metric that leverages the CLIP model \cite{radford2021learning}. 
It calculates the similarity between an image and a generated caption using cosine similarity and a rescaling factor. 
The CIS metric also capitalizes on the capabilities of the CLIP model as its foundation, facilitating the computation of correlations between images and prompts. 
However, while CLIP Score analyzes the correlation of complete prompts, CIS explores the extent to which a generator fully incorporates the components mentioned in the prompt into the generated image.

%Unlike traditional methods that rely on human-written reference captions, CLIPScore assesses caption quality without references. 
%This reference-free approach correlates strongly with human judgments and complements existing metrics. 
%It offers efficient evaluation, making it a valuable tool for assessing image captioning quality. 

% \subsection{Human Related Metric}

% However, the IS has its drawbacks. It fails to capture intra-class diversity, is influenced by the prior distribution over labels, and is sensitive to model parameters and implementations. It also requires a large sample size to produce reliable results. Interestingly, even a simple class-conditional model that memorizes one example per ImageNet class can achieve a high IS, showing its limitations.

% Unbiased FID and IS: " Effectively unbiased fid and inception score and where to find them"
% Must cite
% Vedio fid"FVD: A NEW METRIC FOR VIDEO GENERATION"

% A brief classification of the best solution from the open literature.
% Short description of each relevant solution.
% A detailed criticism of each presented solution, especially in the domains in which the proposed solution is expected to be better.
% Both of them utilized a pretrained InceptionNet \cite{szegedy2016rethinking} as a based to evaluate the  

\section{Inequality of Multi-Component Generation}

We first define the problem of inequality when generative models are tasked to generate images comprising multiple components. 
Consider a case where there is a prompt $T_K$ containing $K$ components. 
We assume that $K>0$ as the prompt contains at least one component. 
With these conditions, we define a generator function $G_{\theta}$ parameterized by $\theta$, which takes a prompt $T_K$ and produces an image, $G_{\theta}:T_K \rightarrow I$. 
The function $F$ then measures the component counts present in the generated image, mapping the image and the prompt to a value representing the number of successfully generated components. 
The relationship between these functions and the problem can be formalized by:

\begin{equation*}
    F(G_{\theta}(T_K), T_K) < K \quad \text{for } K \gg 1
\end{equation*}

Such inequality defined that the generators do not include all components from the original prompt, especially when $K \gg 1$. 
This suggests there exists a gap between the capabilities of existing text-to-image generators and the ideal case where all components in the prompt are accurately represented in the generated image, or $F(G_{\theta}(T_K), T_K) = K$, given a complete generator with a ‘perfect’ $\theta$. 
The formulation assumes the existence of a function $F$ that can accurately measure the number of components in an image, a function that is introduced as part of an evaluation approach in this paper. 
However, such formulation did not consider the quality of the generated image itself; 
Yet, it can be indirectly inferred by the quality of the image affecting the identification of components by $F$.

\section{Components Inclusion Score (CIS)}

The Components Inclusion Score (CIS) is a quantitative metric designed to provide a ratio of how completely a generator incorporates mentioned components from the prompt into the generated image.
% It provides a ratio of how completely a generator incorporates mentioned components from the caption into the generated image.
Ideally, for each component in the prompt, the system should render an equivalent visual feature in the image. 
For example, the given prompt is 
\textit{`A photo of an acoustic guitar, a balaclava ski mask, a sock, and a vase.'}; 
Therefore, the generator should generate images with all the components (acoustic guitar, balaclava ski mask, sock, vase) included in the prompt.
The score is then computed as a normalized sum of these successfully incorporated components.
Thus, the higher the CIS, the more capable the model is of generating complex images composed of many components. 
As a side note, we used `a photo of' as a prefix for the prompts when combining the components \cite{radford2021learning}, but it is pointed out that other prefix alternatives could also be effectively utilized \cite{hessel2021clipscore}.

To begin, we produce $M$ prompts, with each prompt containing $K$ components sampled from a labels set, $\left \{ T_{K,1}, T_{K,2}, T_{K,3}, \cdots, T_{K,M} \right \} \sim  Labels$.
In this work, we use the ImageNet labels as the labels set, $Labels = \left \{c_1, c_2, \cdots, c_P\right \}$, with $c$ being the component.
Then, for each prompt $T_{K,j}$, the image generator $G$ generates $N$ images $G(T_{K,j}) = \left \{I_1, I_2, I_3, \cdots , I_N \right \}$.
In parallel, a lookup table denoted as $L$ is constructed from all the components in $T_{K,j}$, producing a collection of prompts $V_j$. 
This collection includes all possible combinations of components from $T_{K,j}$ as well as an empty string indicator to handle cases where no required components exist in the image.
% TODO the usage of lookup table
The softmax probability $\hat{p}_{i,j}$ is computed by the CLIP model for each generated image $I_i$ corresponding to the prompts in $V_j$, denoted as $\hat{p}_{i,j} = CLIP(I_i, V_j)$.
Specifically, we used the \textit{ViT-B/32} version model to evaluate the correlation between the generated image with the prompts from the lookup table.
This model uses a Vision Transformer with 12 transformer layers and 86M parameters to compute the cosine similarity of corresponding image-caption pairs, in our case, $I$ and $V$ \cite{radford2021learning}.
Then an individual score $S_{i,j}$ is computed as a normalized sum of these successfully incorporated components from the prompt into one generated image.

\begin{equation}
    S_{i,j} = \frac{L(argmax(\hat{p}_{i,j}))}{K}
\label{eq1}
\end{equation}

where function $L(argmax(\hat{p}_{i,j}))$ finds the number of components successfully identified from the lookup table $L$ with $\hat{p}_{i,j}$.
The process of Eq.~\ref{eq1} is repeated for all the images generated.
Finally, the metric $CIS_K \in [0,1]$ with the number of components $K$ is calculated as:

\begin{equation*}
    CIS_K = \frac{\sum^N_i \sum^M_j S_{i,j}}{N M}
\end{equation*}

The framework of the CIS is illustrated in \Figref{fig1}. 
The metric in its spirit measures the capability of an image generator to effectively generate and incorporate multiple components from the prompts;
However, this metric does not consider the aesthetic quality of the generated images.
Yet, the quality of such images can be gauged by the accuracy with which the individual components can be identified by the CLIP model.

% Philosophical essence of the proposed solution.
% Why the proposed solution is without drawbacks of existing solution(s).
% What is the best methodology to prove the superiority of the proposed solution, and under what conditions that holds.

\section{Multi-Component Image Dataset (MCID)}

% TODO not the only reason to create MCID
To effectively evaluate the validity of CIS and for use in fine-tuning tasks, we constructed the Multi-Component Image Dataset (MCID). 
Each entry in the MCID consists of a multi-component image and its corresponding prompt that serves as the ground truth for the visual elements present in the image. 
The number of components present in these images ranges from 1 to 8 and the dataset consists of 160k multi-component images for each component, 1.28M images in total for the whole dataset.

\begin{figure}[t]
\centering
\includegraphics[width=0.9\columnwidth]{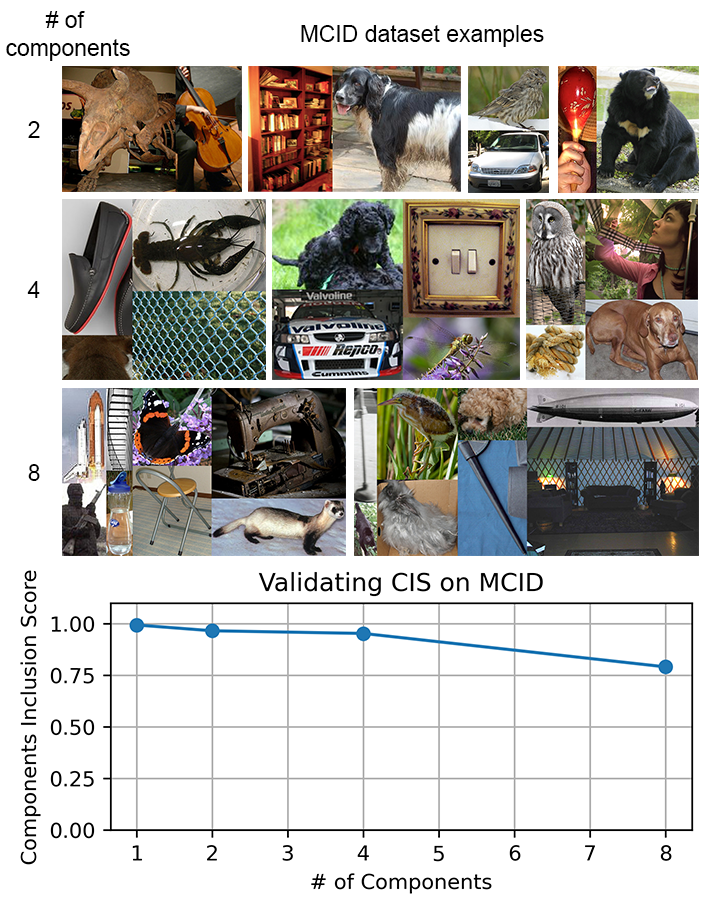}
\caption{
\textbf{Multi-Component Image Dataset (MCID) is used to validate the CIS metric.} 
\figtop Examples of the MCID dataset comprise images combined with multiple components. 
\figbottom The performance of the CLIP model shows near-optimal CIS values in images from MCID. 
Despite there being a decline in performance as the number of components increases due to the limitation of the CLIP model, the outcomes remain to affirm the role of CIS as a metric for benchmarking image generators' capability to incorporate multiple components in a single image.
}
\label{fig2}
\end{figure}

The dataset curation process began with creating a list of prompts, each containing a number of components in it.
The components are sampled from the ImageNet labels.
Following that, we randomly selected and combined images from ImageNet that were presented in the prompt, to encompass a variety of subjects and compositions across images.
Images of varying shapes are combined in such a way that the resulting resolution is as close to a square as possible.
Some samples of the dataset are shown in \Figref{fig2}.

% Experiment that this metric is legit
\subsection{Assessment of CIS Metric Quality based on MCID}

% Experiment

In this experiment, we established a benchmark that enables us to quantify the effectiveness of the CIS in accurately assessing the models' ability to integrate multiple components.
We evaluated the validity of the CIS using images with intentionally incorporated components. 
For this reason, we used the Multi-Component Image Dataset, where the components in the image match the given prompt.
The assessment proceeds by calculating the CIS for the image distributions across $1$, $2$, $4$, and $8$ component counts. 
The different component counts help analyze the effectiveness of each model incorporating an increasing number of visual representations with varying levels of complexity in the input prompt. 
Thus, the scores $CIS_1, CIS_2, CIS_4, CIS_8$ are the factor levels for the number of components in a prompt that the model must capture. 
For instance, $CIS_1$ could denote a single component, $CIS_2$ two components, and vice versa. 
Given the deliberate design of the dataset, the optimal CIS value we hypothesized to observe is a value close to 1, indicating the successful identification of all visual components.

The result in \Figref{fig1} shows that in the image distribution with a lesser number of components, the CIS values indicated near-optimal performance, reflecting the image distribution's successful integration of the visual components within the images.
However, the CIS showed a downward trend with the increasing number of components per image.
This trend signifies that as the images incorporated more visual components, the CLIP model found it progressively more challenging to accurately identify all these components from the image.
This highlights that CLIP Model acts as a bottleneck in identifying a large number of components in the image.
It is important to note that the CLIP model is not specifically trained from images with a high number of distinct components simultaneously.
This result, while slightly lower than anticipated given the current capabilities of the CLIP model, still shows that CIS as a metric can be effectively used to evaluate and compare image generation models under a common baseline.
% still solidifies the CIS's role as a metric in evaluating and comparing image generation models.

%presenting a solid framework for performance evaluation of generative models

\section{Evaluating Image Generation Models}

%This result, while being slightly lower than anticipated, is still acceptable given the current capabilities of the CLIP model. 

%In this experiment, we establish a benchmark that enables us to quantify the effectiveness of the CIS in accurately assessing the models' ability to integrate multiple visual components.
%We evaluate the validity of the CIS in discerning between images with intentionally incorporated components and those without. 
%The former comprises our Multi-Component Image Dataset, with the components in the image matching the given prompt.
%The latter consists of randomly chosen images, sampled from the ImageNet dataset.

%The assessment proceeds by calculating the CIS for the two image distributions across 1, 2, 4, and 8 component categories. 
%For the Multi-Component Image Dataset, we hypothesize an optimal CIS of close to 1, indicating the successful identification of all visual components. 
%In contrast, we anticipate a CIS of close to 0 for the set of random images, implying the absence of specific components. 

% Result of the experiment

% Experiment that evaluates the models
\subsection{Experimental Setup}

\begin{figure*}[!t]
\centering
\includegraphics[width=1\textwidth]{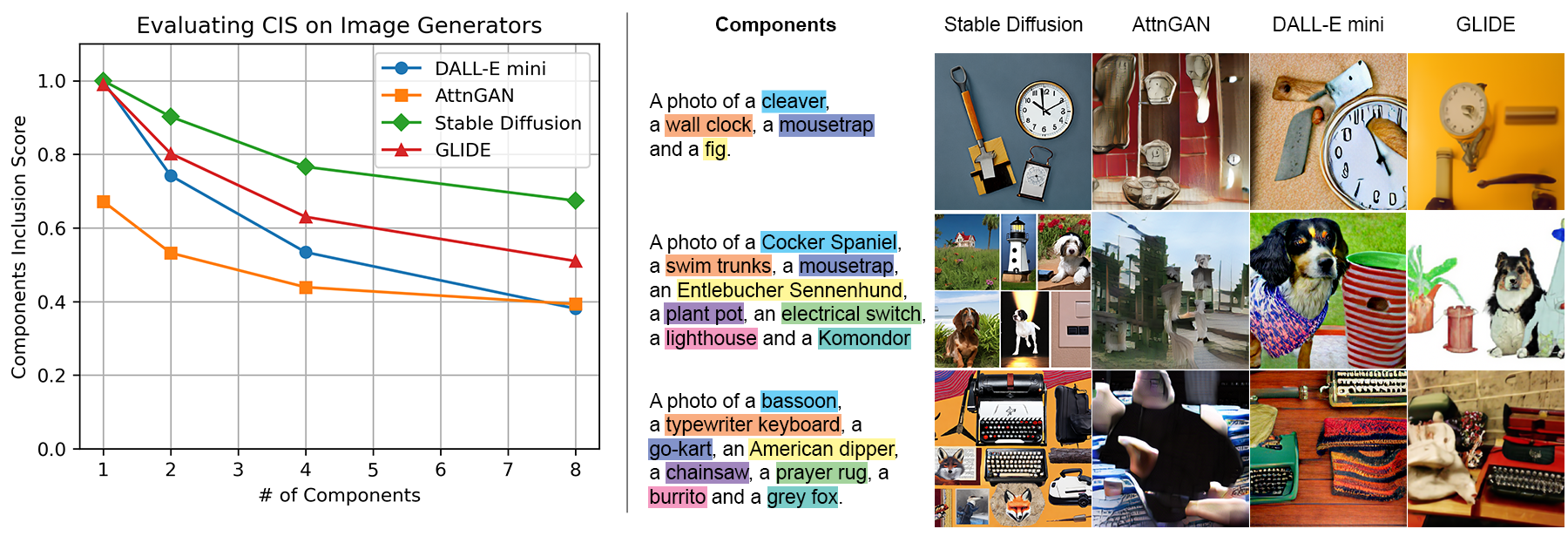} % Reduce the figure size so that it is slightly narrower than the column.
\caption{
\textbf{Image generators have difficulty incorporating all the components mentioned in the prompt.} 
\figleft CIS decreases as the number of components increases, indicating that the evaluated models struggle when tasked with generating images with multiple components. 
Among the models, Stable Diffusion exhibited the most robust performance ($CIS_8 = 0.674$), being able to generate 5 components on average when $K = 8$. 
\figright Some samples of the image generated with multi-components. 
The examples showcase the diminishing image quality and component loss as we increase the number of components in the prompt. 
Notice the degradation of object fidelity and the absence of specified components in the generated images. 
Interestingly, Stable Diffusion can create borders in the images to segregate the components.
}
\label{fig3}
\end{figure*}

\begin{table*}[!t]
\caption{
\textbf{Comparison of Inception Score (IS) and Frechet Inception Distance (FID)}. 
The table shows IS and FID scores across different models with varying numbers of components ($K={1,2,4,8}$). 
It shows a general trend that, as the number of components increases, IS decreases (larger is better), while FID increases (smaller is better).
Note that for FID, the generated image distribution is compared with the MCID with the corresponding number of components.
}
\begin{center}
\resizebox{\textwidth}{!}{
\begin{threeparttable}
\begin{tabular}{l|rrrr|rrrr}
\toprule
\multirow{2}{*}{Image Generation Model} & \multicolumn{4}{c|}{IS $(\uparrow)$} & \multicolumn{4}{c}{FID $(\downarrow)$} \\
 & $K=1$  & $K=2$ & $K=4$ & $K=8$ & $K=1$  & $K=2$  & $K=4$ & $K=8$ \\
\midrule
Stable Diffusion V2 \cite{rombach2021highresolution} & $147.81{\pm2.94}$ &  $72.76{\pm2.54}$ & $35.23{\pm0.98}$ & $18.20{\pm0.30}$ &  $21.03$ & $20.85$ & $23.74$ & $44.29$\\
Dall-E mini \cite{dalle_mini}      & $250.11{\pm3.85}$ & $162.35{\pm4.23}$ & $88.03{\pm1.62}$ & $57.26{\pm1.87}$ &  $20.61$ & $22.62$ & $28.83$ & $64.38$\\
AttnGAN \cite{xu2018attngan}          &  $15.03{\pm0.51}$ &  $17.02{\pm0.38}$ & $15.15{\pm0.50}$ & $12.57{\pm0.23}$ & $106.48$ & $73.68$ & $68.02$ & $67.16$\\
GLIDE \cite{nichol2021glide}            &  $79.90{\pm2.72}$ &  $40.55{\pm1.10}$ & $25.08{\pm0.71}$ & $16.99{\pm0.40}$ &  $21.80$ & $25.43$ & $34.92$ & $57.37$\\  
\bottomrule
\end{tabular}
%\begin{tablenotes}
%\item[a] Note that when comparing images generated by AttnGAN to corresponding components in MCID, the complexity of MCID increases as the number of components increases, leading to an unexpected improvement in FID.
%\end{tablenotes}
\end{threeparttable}
}
\label{tab1}
\end{center}
\end{table*}
% Experiment

In our experiment, we evaluate a set of generative models (DALL-E mini \cite{dalle_mini}, GLIDE \cite{nichol2021glide}, AttnGAN \cite{xu2018attngan}, and Stable Diffusion V2 \cite{rombach2021highresolution}) using the CIS. The evaluation is run across different numbers of components ($K=1,2,4,8$) for each model. 
We sample a large number of prompts ($M=10,000$) across the different settings of $K$. 
For each prompt, the image generator produces $N=16$ images. 

This sample size ensures a robust estimate of the performance metrics \cite{chong2020effectively} and helps mitigate the potential influence of word bias, with the over or under-representation of certain words being more `prominent'.
To investigate if the quality of the generated image also drops as the number of components increases, we computed both the Inception Score (IS), which evaluates the diversity and the quality of the image) and the Frechet Inception Distance (FID), which evaluates the quality of generated images by comparing them with real images. 
The IS and FID are calculated from $30,000$ generated images sampled from each model and component, with the MCID as the real images.

\subsection{Evaluation using Existing Metrics}

The result in \Tableref{tab1} shows that IS decreases (larger is better), while FID increases (smaller is better) as the number of components increases. 
Across all the evaluated models, there is a $15.91\%$ decrease in \IS~ and $9.62\%$ increase in \FID~ ($K=1,2,4,8$).
This suggests that as the component counts increase, there is an inverse relationship with the diversity, quality, and statistical similarity between the distribution of generated images and the `ground truth', MCID.

Through visual inspection, we observed a decline in the visual realism of the generated images as the number of components increased as suggested by IS and FID metrics. 
Components within the images often appear twisted or distorted, and sometimes components are merged in an incoherent manner. 
In most instances, the components were generated incompletely, contributing to the overall degradation in image quality. 
This suggests that complex prompts with multiple components limit the capability of generating high-fidelity images by the current image generation models. 
However, current metrics focus on the quality of the image generated rather than evaluating the accuracy of the image generation models in rendering the image according to the prompt. 

% Result of the experiment
\subsection{Evaluation using CIS Metric}
% TODO I saw people using 'correctness of object locations', what about 'correctness of component generated in the image'?
Using the CIS metric, we now evaluate the correctness of the image generated, i.e., how many components are actually rendered in the image when given in a complex prompt.
The empirical results show a notable decrease in the CIS as the number of components $K$ increases, observed across all the evaluated generative models.
This suggests that as the complexity of the prompt increases, the image generation models fail to render the given prompt accurately. 

Out of all the tested models, the Stable Diffusion model exhibited the most robust performance among the models. 
Despite a slight reduction in CIS as the complexity increased, the decline was less pronounced compared to the other models.
This suggests a greater capability in processing more complex, multi-component prompts.
Interestingly, \Figref{fig3} shows that the model can create borders when generating images with multiple components to segregate the components.
However, a lower CIS score indicates that the Stable Diffusion model, while capable of creating MCID-like images, is not yet perfect in accurately rendering all components in the prompt.

GLIDE and Dall-E mini's exhibited a substantial drop in CIS metric score with each increasing level of prompt complexity, albeit starting from a perfect score when generating images with one component. 
The decrease indicates that a model capable of generating images with a single component may struggle when tasked with generating images with multiple components.
Similarly, the AttnGAN model showed a similar trend but started from a lower baseline showing the superiority of diffusion models over GANs for correctly rendering the components. 
On average, across all the models, the CIS drops $8.53\%$ per component, when $1 \leq K \leq 8$.

To summarize the experiment, while most of the evaluated models perform well in creating a visual representation from a single-component prompt, they encounter challenges with increasing components in prompts.
As the number of components increases, there is a visible drop in the quality of the image generated as quantified by existing metrics IS and FID. 
While the CIS Metric also observes a drop in accurately rendering all the components in the prompt. 
% TODO sk
We note that the drop in the quality of images generated is attributed to rendering multiple components in the image, highlighting that the current image generation models cannot render multiple components, if exist, in a prompt and also maintain a high visual quality of the image. 

\section{Data Distribution Expansion with Multi-Component Images for Improved CIS}

% background
In response to the limitations identified in the previous section, we hypothesized that a training data distribution that includes images with a higher component count could improve the model's ability to generate multi-component images. 
To test this, we fine-tuned the Stable Diffusion model, our best-performing model from the evaluation, using MCID. 
This aimed to determine the impact of a more diverse data distribution on the model's performance.

% experiment
For the fine-tuning process, we employed Low-Rank Adaptation (LoRA) \cite{hu2021lora} as a fine-tuning scheme, adding bottleneck branches to the attention layers of the Stable Diffusion V2. 
We selected a bottleneck rank $r=4$ following the original LoRA paper \cite{hu2021lora} for the best-performing hyperparameter.
The optimization was carried out using the AdamW optimizer, set at a learning rate of $1 \times 10^{-5} $ and a weight decay of $1 \times 10^{-2} $. 
The model was fine-tuned on the subset of MCID of 640k images (8 images for each prompt, for all 1 to 8 components with 10k total prompts).
The results are reported using both CIS and IS metrics; particularly, the IS is used to confirm that high-quality and diverse images are produced, regardless of whether the CIS increases or decreases.

\begin{table}[!t]
\begin{center}
\caption{\textbf{Performance evaluation of the Fine-Tuned and Vanilla Stable Diffusion V2 on MCID, showing CIS and IS across different component counts.} This table presents the CIS and IS values for varying numbers of components ($K = 1, 2, 4, 8)$. Despite a notable decrease in both CIS and IS as $K$ increases, it is important to highlight that the fine-tuned model still outperforms its vanilla counterpart.}
\begin{tabular}{c|rr|rr}
\toprule
& \multicolumn{2}{c|}{Fine-Tuned Model} & \multicolumn{2}{c}{Vanilla Model} \\
$K$ & $CIS (\uparrow)$ & $IS (\uparrow)$ & $CIS (\uparrow)$ & $IS (\uparrow)$ \\
\midrule
1 & $1$ &  $176.42{\pm3.34}$ & $1$ &  $147.81{\pm2.94}$ \\
2 & $0.92$ & $85.38{\pm2.19}$ & $0.90$ & $72.76{\pm2.54}$ \\
4 & $0.86$ & $40.62{\pm1.07}$ & $0.76$ & $35.23{\pm0.98}$ \\
8 & $0.73$ & $21.08{\pm0.29}$ & $0.67$ & $18.20{\pm0.30}$ \\
\bottomrule
\end{tabular}
\label{tab3}
\end{center}
\end{table}

\begin{figure}[!t]
\centering
\includegraphics[width=0.9\columnwidth]{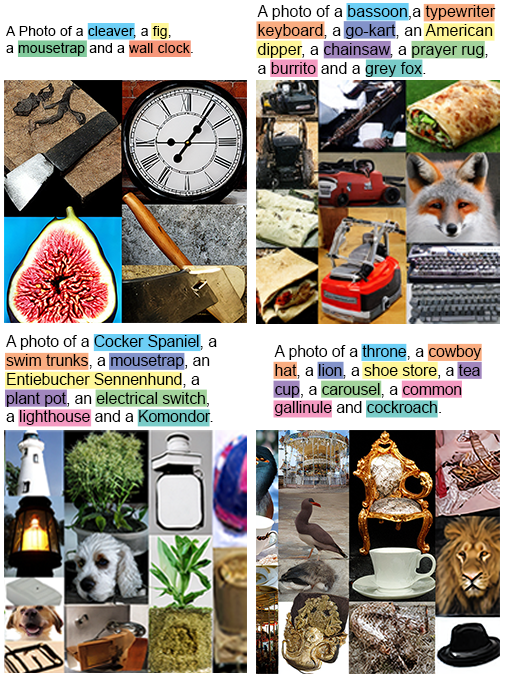}
\caption{
\textbf{Examples of images generated by the Stable Diffusion model fine-tuned with MCID.} 
Similar to MCID, the images display borders to segregate components, yet they can appear distorted with limited natural interactions among elements as the number of components increases.
}
\label{fine_tuned_result}
\end{figure}

% result
The result in \Tableref{tab3} indicates that the fine-tuned model is better than the vanilla counterpart ($4.55\%$ increase overall, when $1 \leq K \leq 8$), and approaching the CIS evaluated on MCID (\Figref{fig2}).
However, it should be noted that the fine-tuning process did not always represent all components' co-occurrences found in the training data. 
Such discrepancy in co-occurrence representation may account for the sub-optimal CIS scores.
Moreover, the comparison of IS in \Tableref{tab3} indicates that while CIS has increased, the fine-tuned model continues to produce high-quality and diverse images.
In addition, visual inspection (\Figref{fine_tuned_result}) of the images generated by the fine-tuned model revealed a tendency to create borders in the images to segregate the components. 

% discussion
%however, hypothetically, a prompt can consist far more than 8 components.
%Model should learn to understand the task and incorporate them, without relying on data distribution.
%Model should learn to take the initiative to place every component in image, requested by the prompt.
% This give a direction that, data distribution should include data with more component in one image. 
%This further opens up a direction for future research to improve models' capability of handling complex multi-component prompts, to render more sophisticated images. 
% include direction to solve this problem
%Note that the overall performance across different models could also be influenced by other factors such as the underlying architecture, loss function (regularization term that punish prompt that do not exist) and distribution of the training dataset.

While the improved CIS is encouraging, we must consider that a model should not solely rely on existing data distributions but should also develop an innate understanding to construct causal relations between objects that have not been seen together in the training data.
Additionally, we acknowledge that overall performance could also be influenced by other factors, including the underlying neural network architecture and the specifics of the loss function (such as regularization terms penalizing unrepresented prompts)
For now, an immediate solution is to include data distribution that involves multi-components, especially those whose co-occurrence is low, while working on enhancing the model architecture and learning methods.

\section{Extended Analysis}

\subsection{Sequence invariant to the Order of Components Presented}

% Experiment

% Especially, the attention model, where components in the middle of the input caption are usually given less attention [cite long attention paper]. 
% To evaluate whether the image generators are sequence invariant to the order of components within the input prompt, we randomized the order of components in the prompt and assessed if the components being generated remain consistent. 
Here, we evaluated the image generators' sequence invariance to the order of components within the prompt by randomizing component order in prompts and assessed the consistency of generated components.
We generated two image distributions with Stable Diffusion ($K=8$, $M=1000$, $N=16$) based on two sets of prompts: the original prompts and the shuffled prompts (randomly rearranged based on the original).
Subsequently, the CLIP model is used to identify the components in the images and verify the presence or absence of each component.
The Chi-squared test for independence is utilized to compare the distribution of generated components between the two groups of images.
Our null hypothesis assumes that no difference in the distribution of detected components in images generated from shuffled versus original prompts. 
Any rejection of the null hypothesis ($p < 0.05$) would signify a significant effect of component sequence on image generation. 

\begin{table}[!h]
\begin{center}
\caption{Comparison of $CIS_8$ of Stable Diffusion under the setting of $K=8$, $M=1000$, $N=16$.}
\begin{tabular}{l|r}
\toprule
Set & $CIS_8 (\uparrow)$ \\
\midrule
Original & 0.664  \\
Shuffled     & 0.678   \\
\bottomrule
\end{tabular}
\label{tab2}
\end{center}
\end{table}

% Result of the experiment
\Tableref{tab2} shows the $CIS_8$ of the original and shuffled set.
The test failed to reject the null hypothesis, indicating that there is no significant effect of component sequence on image generation 
$X^2 (996, N=175918)=1006.76$, $p= .399$.
Consequently, we do not have to shuffle the components in the prompts when generating the image distribution, as it does not affect the stability of CIS, at least when $K \leq 8$.

\subsection{Individual Component Analysis}

Here, we perform an analysis to investigate whether certain components can be `prominent,' meaning they are more likely to be generated, while other components may tend to be ignored. 
The analysis is performed across all evaluated models, but only on components with $K=2,4,8$.

\begin{figure}[!t]
\centering
\includegraphics[width=0.90\columnwidth]{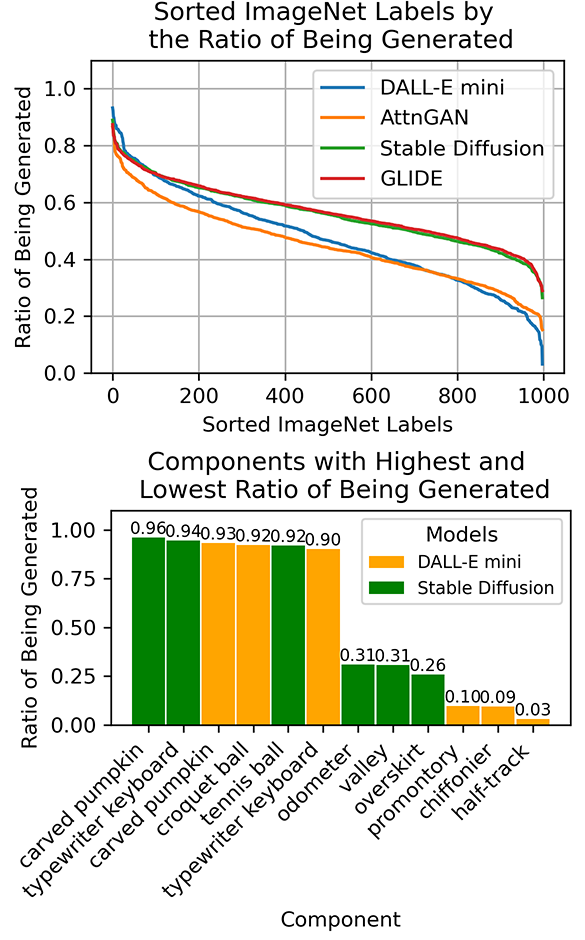}
\caption{\textbf{Image generators may be biased towards certain components.} \figtop The ImageNet labels are sorted by the ratio of components being successfully generated ($K=2,4,8$). This suggests that models have an internal bias to prefer some components over others in a multi-component prompt. \figbottom Some samples showcase components with the highest and lowest ratios of being generated ($K=2,4,8$). 
% TODO WEAK ASS EXPLANATION
Interestingly, there is an overlap between the components, indicating that perhaps some objects are easier to generate and identified or the training dataset is biased.}
\label{fig4}
\end{figure}

\Figref{fig4} (Top) shows the distribution of components based on their generation rate by the corresponding image generation model. 
We observe that some components have a higher probability of getting generated while some components are less likely to be generated in a multi-component prompt.
This suggests that inherently, models have an internal bias to prefer some components over others in a multi-component prompt. 
This bias is more associated with DALL-E mini and AttnGAN models as they have lesser generation rate for most of the components compared to GLIDE and Stable Diffusion. 
Particularly, DALL-E mini exhibits a stronger bias, with certain components having the highest ratio of being generated, but this ratio drastically drops towards the end of the quantile.

\section{Limitations and future work}

% Limitation
The accuracy of CIS appears to be constrained by the limitations of the underlying evaluating model, CLIP. 
As such, the evaluation module within our framework is designed to be replaceable, if a more advanced model or methodology for identifying components in images becomes available. 

Additionally, the MCID combines various images directly, it does not take into account more natural interactions, such as components interacting with each other or sharing the same background. 
This situation also highlights the problem that existing datasets rarely include multiple components within a single image, especially components that would not naturally appear together in real-world scenarios. 
As a result, models need to learn the correlations and interactions between these components on their own, without guidance from more representative training data.

% Future work
We limited our testing to $8$ components as the models appeared to reach their limitations at that point, making further analysis with a greater number of components unlikely to provide additional insights.
However, evaluating a model with more components remains a desirable goal when more capable models become available.
Since the problem is raised, our future works should focus on addressing it from four aspects: (1) enhancing the training data distribution with multiple components, developing (2) a network architecture that adeptly combines multiple components, (3)  technique that applies generative image inpainting iteratively for each component \cite{zeng2020high, zeng2021cr}, and (4) a loss function that penalizes the absent of components. 
% Each of these challenges is challenging on its own, yet overcoming them will be rewarding. 

\section{Conclusion}
% Summary
We introduced the Components Inclusion Score (CIS) as a metric to evaluate image generators' ability to incorporate multiple components within an image. 
Tasks that previously relied solely on human evaluation are now automated through this evaluation framework. 
% We also performed extensive experiments and analyses to validate the metric.
% The metric is validated through a series of experiments and analyses.
The result of evaluating the modern image generators also hinted at the challenges in comprehending spatial correlation, coherence between multiple components, and the integration of these components into a cohesive image.
Guided by this, we have outlined potential future work, the success of which will bring us closer to a model that more accurately approximates human-level understanding in comprehending the objectives of the prompts.

{
    \small
    \bibliographystyle{ieeenat_fullname}
    \bibliography{main}
}


\end{document}